\def\year{2022}\relax
\documentclass[letterpaper]{article} 
\usepackage{aaai22}  
\usepackage{times}  
\usepackage{helvet}  
\usepackage{courier}  
\usepackage[hyphens]{url}  
\usepackage{graphicx} 
\usepackage{natbib}  
\usepackage{caption} 
\DeclareCaptionStyle{ruled}{labelfont=normalfont,labelsep=colon,strut=off} 
\makeatletter  
\newif\if@restonecol  
\makeatother

\usepackage[ruled,linesnumbered]{algorithm2e}

\usepackage{float}
\usepackage{newfloat}
\usepackage{listings}
\floatstyle{ruled}
\newfloat{listing}{tb}{lst}{}
\floatname{listing}{Listing}
\title{Graph Pointer Neural Networks}

\author{
    Tianmeng Yang\textsuperscript{\rm 1,2}\thanks{This work is conducted during the author's internship at Microsoft Research Asia.}\equalcontrib,
    Yujing Wang\textsuperscript{\rm 1,2}\equalcontrib,
    Zhihan Yue\textsuperscript{\rm 1},
    Yaming Yang\textsuperscript{\rm 2},
    Yunhai Tong\textsuperscript{\rm 1},
    Jing Bai\textsuperscript{\rm 2}
}
\affiliations{
    \textsuperscript{\rm 1}School of Electronics Engineering and Computer Science, Peking University\\
    \textsuperscript{\rm 2}Microsoft Research Asia\\
    \{youngtimmy, zhihan.yue, yhtong\}@pku.edu.cn,
    \{yujwang, yayaming, jbai\}@microsoft.com
}

\usepackage{bibentry}
\usepackage{booktabs}
\usepackage{amssymb}
\usepackage{xcolor}
\usepackage{subcaption}
\usepackage{amsmath}
\usepackage{hyperref}

\begin{document}
\maketitle

\begin{abstract}
Graph Neural Networks (GNNs) have shown advantages in various graph-based applications. Most existing GNNs assume strong homophily of graph structure and apply permutation-invariant local aggregation of neighbors to learn a representation for each node. However, they fail to generalize to heterophilic graphs, where most neighboring nodes have different labels or features, and the relevant nodes are distant. Few recent studies attempt to address this problem by combining multiple hops of hidden representations of central nodes (i.e., \textit{multi-hop-based approaches}) or sorting the neighboring nodes based on attention scores (i.e., \textit{ranking-based approaches}). As a result, these approaches have some apparent limitations. On the one hand, multi-hop-based approaches do not explicitly distinguish relevant nodes from a large number of multi-hop neighborhoods, leading to a severe over-smoothing problem. On the other hand, ranking-based models do not joint-optimize node ranking with end tasks and result in sub-optimal solutions. In this work, we present Graph Pointer Neural Networks (GPNN) to tackle the challenges mentioned above. We leverage a pointer network to select the most relevant nodes from a large amount of multi-hop neighborhoods, which constructs an ordered sequence according to the relationship with the central node. 1D convolution is then applied to extract high-level features from the node sequence. The pointer-network-based ranker in GPNN is joint-optimized with other parts in an end-to-end manner. Extensive experiments are conducted on six public node classification datasets with heterophilic graphs. The results show that GPNN significantly improves the classification performance of state-of-the-art methods. In addition, analyses also reveal the privilege of the proposed GPNN in filtering out irrelevant neighbors and reducing over-smoothing.  

\end{abstract}

\section{Introduction}
\begin{figure}[t]
	\centering
	\begin{subfigure}[t]{0.45\columnwidth}
         \centering
         \includegraphics[width=\columnwidth]{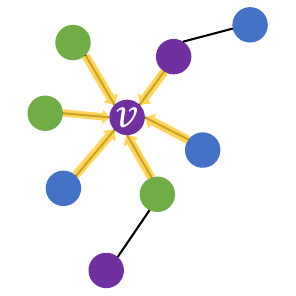}
         \caption{Local aggregation.}
     \end{subfigure}
     \hfill
     \begin{subfigure}[t]{0.45\columnwidth}
         \centering
         \includegraphics[width=\columnwidth]{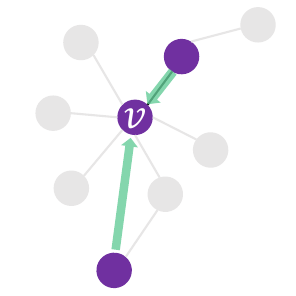}
         \caption{Non-local aggregation.}
     \end{subfigure}
       \caption{ Visual illustration of the local aggregation of existing GNNs and non-local aggregation in GPNN in one propagation step. The colors of nodes represent their labels, while grey means ignored. Traditional GNNs aggregate all nodes in the local neighborhood including noises, while GPNN selectively filters the irrelevant nodes and captures non-local features.}
       \label{fig:aggregate}
\end{figure}

Graph Neural Networks (GNNs) have shown advantages in various graph-based applications. Most existing GNNs assume strong homophily of connected nodes have been successfully applied to representation learning on graphs, as well as multiple real-world applications from web-scale recommendation~\cite{ying2018graph} to molecular chemistry inference~\cite{gilmer2017neural}. Most existing approaches are based on a framework of message-passing neural networks (MPNNs), including ChebyNet~\cite{chebynet}, GCN~\cite{kipf2016semi}, GAT~\cite{velivckovic2017graph} and GIN~\cite{xu2018powerful}. They learn each node’s representation by aggregating feature information from its neighbors. The aggregation often needs to be permutation-invariant (e.g., Mean, Max or Sum) as there is no ordering information of the neighboring nodes. These methods are effective with an assumption of strong homophily for graphs (such as citation networks~\cite{newman2002assortative}), where neighboring nodes always possess similar features and belong to the same class. However, this inductive bias is not suitable for graphs with heterophily. In heterophilic graphs, most connected nodes are dissimilar and may belong to different classes, while the semantically relevant nodes are often multi-hops away. As Figure~\ref{fig:aggregate} (a) shows, normal local aggregation in GNNs might introduce noises for node representation in heterophilic graphs. In such scenarios, even models that ignore the graph structure altogether (e.g., MLPs) can outperform state-of-the-art GNNs~\cite{zhu2020beyond}.

Few recent studies address heterophilic graphs by combining multiple hops of hidden representations of central nodes or ranking the nodes with attentions scores. For example,~\citet{zhu2020beyond} combine each node's ego-features and intermediate representations of multiple hops of neighbors to boost learning from the heterophilic graph structure; \citet{yuan2021node2seq} compute attention scores between neighboring nodes and drop the nodes with lower scores manually. The limitations of these methods lie in two aspects. First, multi-hop-based approaches do not explicitly distinguish relevant nodes from a large number of multi-hop neighborhoods. The noises are still mixed up with helpful information and lead to severe over-fitting and over-smoothing issues. Second, the ranking procedure in state-of-the-art solution~\cite{yuan2021node2seq} is non-differentiable, e.g., it cannot be optimized jointly with the target classification tasks and leads to sub-optimal results. GAT~\cite{velivckovic2017graph} is able to assign different weighting scores to each neighbor, but it still suffers from over-smoothing and cannot fundamentally solve the limitation when the number of neighbors or hops are extremely large~\cite{chen2020measuring}.

In this paper, we tackle the challenges above by proposing a novel Graph Pointer Neural Networks (PGNN). For each node, we first sample a sequence of nodes from the local and remote neighborhoods. A pointer network is then utilized to select the most relevant and valuable neighboring nodes, which constructs a new sequence ranked by relations to the central node. Afterward, a 1D-convolutional layer is applied to the node sequence to capture high-level semantic features. As Figure~\ref{fig:aggregate}(b) shows, the central node can directly capture helpful information from a distant node while ignoring the irrelevant ones from the nearest neighborhood. We can easily sample sufficient nodes from multiple-hop neighbors in the implementation, enabling the model to capture long-term dependencies of nodes in a heterophilic graph. 

To prove the effectiveness of GPNN, we conduct extensive experiments on a variety of real-world graphs with heterophily properties, including web-page linking networks and co-occurrence networks. The results show that our methods consistently improve the performance of existing GNNs over all benchmarks, with an average lift of 6.3\% over the best state-of-the-art methods. Thorough analyses also reveal the privilege of the proposed GPNN in filtering out irrelevant neighbors and reducing over-smoothing. 

In summary, the major contributions of this paper are three-fold: 
\begin{itemize}
    \item First, we propose a novel framework termed Graph Pointer Neural Networks (GPNN) tailored for heterophilic graphs. The adoption of a pointer network enables GPNN to distinguish crucial information from distant nodes while filtering out irrelevant or noisy ones in the nearest neighbors. Besides, the 1D-convolutional layer can extract high-level structural information from the ranked sequence of relevant nodes, enriching the representation of nodes in heterophilic graphs.
    \item Second, experimental results show that GPNN consistently outperforms previous methods, with an average lift of 6.3\% over the best state-of-the-art method. Especially, we improve 1.9\% on Chameleon and 3.0\% on Cornell respectively over the second-best results. 
    \item Last but not least, we further demonstrate the privilege of GPNN through extensive analysis. We prove that the ranked sequence produced by the pointer network greatly enhances the homophily property of neighboring nodes. Moreover, GPNN mitigates the over-smoothing problem and performs better than GCN and GAT as the number of layers grows.
\end{itemize}

\section{Related Work}
\paragraph{Graph Neural Networks}
Graph Neural Networks have many variants and applications. Here we focus on a brief introduction of representation learning for graph nodes in a supervised or semi-supervised setting. Most existing approaches follow a message-passing framework and use a permutation-invariant local aggregation scheme to update each node's representation~\cite{scarselli2008gnn,chebynet,kipf2016semi,velivckovic2017graph}. For example, GCNs~\cite{kipf2016semi} average features from each node's directly connected neighbors (including the node's self feature) to update its representation. GATs~\cite{velivckovic2017graph} introduces the attention mechanism~\cite{vaswani2017attention} to attend over all neighbors with the learned weights. Sampling-based techniques have been developed for fast and scalable GNN training, such as GraphSAGE~\cite{hamilton2017inductive} and FastGCN~\cite{chen2018fastgcn}. Simplifying methods \cite{tiezzi2021LP-GNN,wu2019SGC} are also proposed to make the GNN models more easily implementable and more efficient.  Mixhop~\cite{abu2019mixhop} and Graph Diffusion Convolution(GDC)~\cite{klicpera2019diffusion} explored combining feature information from multi-hop neighborhoods, while PPNP~\cite{klicpera2018PPNP} and PPRGo~\cite{bojchevski2020PPRGo} derived an improved propagation scheme of high-order information based on personalized PageRank. In order to directly learn weights in the local filter, LGCN~\cite{gao2018lgcn} adopt regular 1D convolutions through top-k ranking. Besides, data augmentation and consistency regularization are applied in GRAND~\cite{feng2020grand} to increase the robustness of GNN models as well as reduce the risk of over-smoothing. Other recent works \cite{ying2021graphormer,kreuzer2021SAN} generalize standard Transformers \cite{vaswani2017attention} to graphs and preserve the structural information by extra encodings, such as centrality encodings, spatial encodings, Edge Encodings, or Laplacian eigen-vectors as positional encodings. 

\paragraph{GNNs addressed heterophily}
Without the inductive bias of strong homophily, traditional GNNs based on local aggregation face a severe performance reduction on the heterophilic graphs. To address this challenge, Geom-GCN~\cite{pei2020geom} proposes to pre-compute unsupervised node embeddings and defines a new graph convolution with a structure neighborhood built by geometric relationships in the latent space. Furthermore, some other works~\cite{liu2020non,jiang2021gcnsl} share the idea of re-connect graphs to improve the homophily property. Recently, H2GCN~\cite{zhu2020beyond} combines a set of intermediate representations including ego- and high order neighborhoods to boost learning from a heterophilic graph structure. Node2Seq~\cite{yuan2021node2seq} proposes to sort the nodes and apply regular 1D convolutions. These approaches have shown advantages on datasets with heterophilic graphs. However, since the irrelevant features are mixed up, and the ranking procedure does not jointly optimize, these methods fail to generate the optimal results.

\section{Preliminaries}
Let $\mathcal{G=(V,E)}$ be an undirected graph with node set $\mathcal{V}$ and edge set $\mathcal{E}$. The nodes are associated with a feature matrix $X \in \mathbb{R}^{N \times F}$, where $N = |\mathcal{V}|$ denotes the number of nodes and $F$ denotes the number of input features. $A \in \{0,1\}^{N \times N}$ is the adjacency matrix, where $A_{uv}=1$ means the node $u$ and node $v$ are connected. $x_v$ is the feature vector of node $v$, and the corresponding label is $y_v$. The $k$-hop neighborhood of node $v$ is denoted as $N_k(v)$. For example, the directly connected (1-hop) neighborhood is $N_1(v)$, including self-loops. 

\subsection{Homophily and Heterophily}
Graphs such as community networks and citation networks are often of high homophily, where the linked nodes are more likely to have similar features and belong to the same class. However, there are a large number of real-world graphs with heterophily (e.g., web-page linking networks~\cite{ribeiro2017struc2vec}). That is, the linked nodes usually have dissimilar features and belong to different classes. It is worth noting that heterophily is different from heterogeneity, as a heterogeneous network means that the network has multiple types of nodes and different relationships between them. 

To clearly measure the homophily or heterophily of a graph, we follow \citet{pei2020geom} to define the homophily ratio $H(\mathcal{G})$ and use it to distinguish graphs with strong homophily or heterophily:
\begin{equation}
    H(\mathcal{G}) = \frac{1}{|\mathcal{V}|} \sum_{v\in \mathcal{V} } \frac{\sum_{u\in N_1(v)} (y_u=y_v)}{|N_1(v)|}
\end{equation}
A high homophily ratio $H(\mathcal{G}) \to 1 $ means that the graph is with strong homophily while a graph of strong heterophily has a small homophily ratio $H(\mathcal{G}) \to 0$.

\subsection{Traditional GNNs}
\textbf{Message Passing Framework.}
Most existing GNNs adopt message-passing framework, which applies local aggregation to learn node representations. 
At each propagation step $t$, the hidden representation of node $v$ is derived by:
\begin{align}
    h_v^t = f({\rm aggr} (h_u^{t-1} | u \in N_1(v)))
\end{align}
where $h_v^0 = x_v$, $f(\cdot)$ is the transformation function between two propagation steps, and ${\rm aggr}(\cdot)$ aggregates all 1-hop neighbors' features. For example, GCN~\cite{kipf2016semi} aggregates and updates the node features by:
\begin{align}
    h_v^t = \sigma(W\sum_{u \in N_1(v)}\frac{1}{\sqrt{\hat{d}_u\hat{d}_v}} h_u^{t-1}))
    \label{eq:gcn}
\end{align}
where $\sigma$ is the activation function, $W$ is a learnable weight matrix, $\hat{d}_v$ is the degree of node $v$ obtained from the adjacency matrix with self-loops $\hat{A} = A + I$, $1/\sqrt{\hat{d}_u\hat{d}_v}$ denotes the weight between node $u$ and $v$. The propagation depth $t$ is usually limited to prevent the over-smoothing phenomena. However, under the situation of heterophily, most of the directly connected nodes are noisy, while semantically similar nodes are always distant.

\textbf{Attention-based approaches} compute the attention scores between connected nodes as: 
\begin{align}
    \alpha_{uv} = a(Wh_u,Wh_v)
\end{align}
where $a(\cdot)$ is the function computing the similarity between two connected nodes. Based on the attention scores $\alpha_{uv}$, the neighboring nodes can be considered differently or sorted manually.

\textbf{Multi-hop-based approaches} consider the representations of different propagation steps to combine information from different distance as:
\begin{align}
    h_v^t = {\rm combine}(h_v^0,h_v^1,…,h_v^{t-1})
\end{align}
but the noises of the local neighborhood are also considering and always facing the over-smoothing issue. 

\subsection{Sequence-to-Sequence Model}
Formally, given a sequence $s$ containing $L$ nodes embedded, 
\begin{align}
    s  = \{\hat{x}_1,\hat{x}_2,…,\hat{x}_L\} \qquad \in \mathbb{R}^{L \times d}
\end{align} 
our target is to select the top-$m$ most relevant nodes in them and output a sequential list reflecting the appropriate order. We denote the $m$ indices of selected nodes with $c =\{c_1,c_2,…,c_m\}$,  the sequence-to-sequence model aims to compute the conditional probability:
\begin{align}
    p_{\vartheta}(c_i|c_1,c_2,…,c_{i-1},s;\vartheta)
    \label{eq:conditional}
\end{align}
and learn the parameters $\vartheta$ by maximizing the probability:
\begin{align}
    \vartheta^* = \mathop{{\rm arg\ max}}\limits_{\vartheta}\sum\limits_{s,c}\log p(c|s; \vartheta)
\end{align}

\subsection{Motivation of GPNN}
Motivated by the limitations of traditional GNNs, we propose a novel GNN framework termed Graph Pointer Neural Networks (GPNN), which constructs node sequences containing local neighboring nodes and high-order distant nodes to capture both local and non-local semantic information. A pointer network is then leveraged to select and rank the most relevant nodes in structure and semantics. The pointer network can be jointly optimized with graph embedding, resulting in significant improvement over state-of-the-art methods. The details are described in the next section.

\section{Graph Pointer Neural Networks} \label{sec:gpnn}

\begin{figure*}[htbp]
	\centering
        \includegraphics[width= \linewidth]{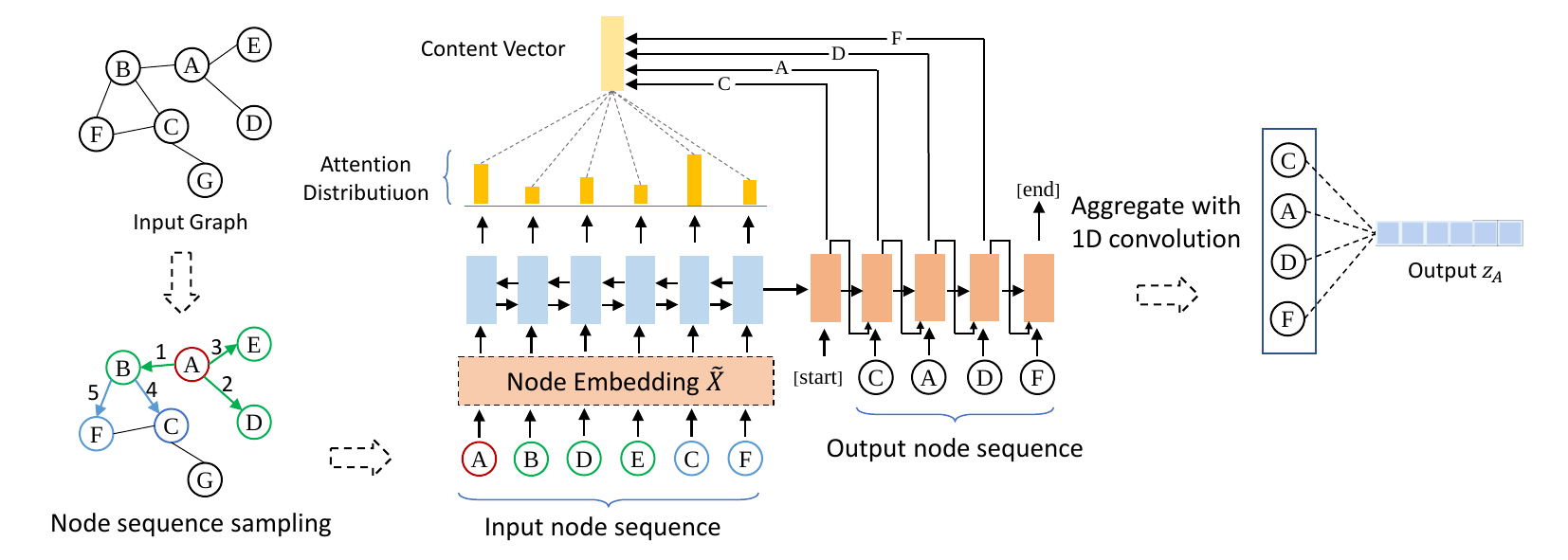}
        \caption{ An illustration of the graph pointer generator layer. With the central node $A$ and a sampling depth $k$=2, the neighbors within two hops are assembled after the node sequence sampling. The pointer network then selects the most relevant nodes to $A$, followed by a 1D-convolution layer to extract high-level and non-local features at the end.}
        \label{fig:archtecture}
\end{figure*}

In this section, we present our proposed Graph Pointer Neural Networks (GPNN) for node representation learning of heterophilic graphs (see Figure \ref{fig:archtecture}). We first describe the node sequence sampling strategy to construct a node sequence including local and high-order neighborhood nodes. Then we introduce the graph pointer generator, which includes a node embedding layer, a pointer network to select and rank these nodes, and a 1D-convolution layer to extract high-level semantics from the ranked sequence. Finally, we introduce the proposed graph pointer neural networks for node classification tasks.

\subsection{Multi-hop node sequence sampling}
\label{sec:sampling}
In graphs with heterophily, nodes with helpful information are located both in the local and non-local high-order neighborhoods. In order to capture these two affinities between nodes, we aim to construct a node sequence for each node that contains neighbors from multiple hops. Unlike the grid-like data such as time series and images, there are several challenges to turn generic graph structure into sequences. The number of node's adjacent neighbors is always varying, and there is no ordering information among them. 
To tackle these challenges, we propose a multi-hop node sampling strategy to construct node sequences for encoding structural and semantic information. Algorithm \ref{alg:algorithm} describes the procedure in detail. For each node in the graph, we sample the nodes from its $1$-hop neighborhood to $k$-hop neighborhood. We use a Breath-First-Search (BFS) to expand adjacent neighbors. Nodes in the sequence are ranked by the distance to the central node.
Theoretically, since the sampling depth hyperparameter $k$ can be set flexibly to cover the whole graph, the strategy is eligible to capture long-range dependencies even the two nodes are distant. In case some nodes have two many neighbors, we set a fixed max length $L$ of the sequence and stop sampling when meeting this limitation.

\begin{algorithm}[h]
\caption{Multi-hop node sequence sampling}
\label{alg:algorithm}
\LinesNumbered 
\KwIn{Adjacency matrix $A$, Number of nodes $N$, Sampling depth $k$}
\KwOut{Node sequence $S$}
\textbf{Inital $S \to \varnothing $}\\
    \For{$node_i = 1,2,…,N$}{
        $S[node_i].append(node_i)$\\
    }
    \For{$i = 1,2,…,k$}{
        $A = A^i$\\
        \For{$node_i = 1,2,…,N$ and $node_j = 1,2,…,N$}{
            \If{$A_{ij}==1$ and $node_j \notin S[node_i]$}
                {$S[node_i].append(node_j)$}
        }
    }
\textbf{Return $S$}
\end{algorithm}

\subsection{Graph Pointer Generator}
The sampled node sequence contains both relevant and irrelevant nodes for the central node. In order to select the most informative nodes out and eliminate noises, we consider the selection as a sequence-to-sequence problem: the nodes in the output sequence are selected from the input sequence, while the resulting order reflects the relevance or relationship with the central node. 

We leverage the Pointer Networks ~\cite{vinyals2015pointer} as an embedded component to achieve this goal. 
For each input sequence of neighboring nodes, we first embed the node feature vector into a latent space of dimension $d$, then LSTMs~\cite{hochreiter1997long} are utilized in the pointer network for the sequence-to-sequence task. 

\subsubsection{Node embedding.} 
Node embedding is a fundamental method to preserve the connection and distance pattern in a graph. In this step, we apply a GCN layer aiming to capture the local structural information of each node. With the input nodes feature $X$, the output embedding is denoted as:
\begin{equation}
    \hat{X} = GCN(X) \qquad \in \mathbb{R}^{N \times d}
    \label{eq:neiborhood}
\end{equation}
where we use the vanilla GCN layer in Equation (\ref{eq:gcn}), and embed the feature vectors into $d$-dimentional hidden representations.

\subsubsection{Pointer Network}
After embedding, the sampling node sequence is fed into a pointer network to select the most relevant nodes and rank the nodes by the relevance or relationship to the central node. 

We adopt a sequence-to-sequence architecture based on LSTMs for the pointer network to model the conditional probability $p_\vartheta$ in Equation (\ref{eq:conditional}). Two separate LSTMs are applied as \textbf{Encoder} and \textbf{Decoder} respectively.

The encoder generates hidden states for the input sequence. At each time step $i$, $\hat{x}_i$ is fed into the encoder, the hidden state is denoted as:
\begin{align}
    e_i = tanh(W[e_{i-1},\hat{x}_i])
\end{align}
where $e_0$ is initialed to $0$. After $L$ time steps, we obtain $L$ hidden states of input sequence and combine them into a content vector $E=\{e_1, e_2, \dotsc, e_L\}$ that records the information of entire sequence of $L$ nodes.

The decoder then selects node with attention scores among the $L$ nodes. At each output time $i$, the hidden state of decoder $d_i$ is: 
\begin{align}
    & d_i = tanh(W[d_{i-1},\hat{x}_{c_{i-1}}])
\end{align}
where $d_0$ is the output hidden state $e_L$ from the encoder, $c_{i-1}$ is the index of selected node at time step $i-1$, $c_0$ is a signal of $[start]$. 
We compute the attention vectors over the input sequence as following:
\begin{align}
    & u_j^i = v^Ttanh(W_1e_j+W_2d_i)\qquad j \in (1,2,…,L) \\
    & p(c_i|c_1,c_2,…,c_{i-1},s) = softmax(u_i)
\end{align}
where $v$, $W_1$, and $W_2$ are learnable parameters of the decoder model, $softmax$ normalizes the vector $u_i$ (of length $L$) to be an output distribution over the $L$ nodes of input sequence.
As Figure \ref{fig:archtecture} shows, with the probability distribution $u_i$ , we can intuitively use $u_j^i$ as pointers to select the $i$-th node of output sequence, until all the top-$m$ nodes are selected step by step.
After $m$ output time steps, we obtain the sequence of top-$m$ relevant nodes
\begin{align}
    o = \{\hat{x}_{c_1},\hat{x}_{c_2},\dotsc,\hat{x}_{c_m}\} \qquad \in \mathbb{R}^{m \times d}
\end{align}
which is ranked with the output order.

\subsubsection{Non-local aggregation}

Based on the ranked sequence output from the graph pointer generator, we extract and aggregate structural and semantic features from non-local neighbors.
A regular 1D-convolution layer is first applied to extract the affinities between the sequential nodes even the nodes are distant in the generic graph. Let the output channel be $d'$, we then aggregate all the node features to obtain a representation $z \in \mathbb{R}^{1 \times d'}$ of the central node.

\begin{align}
    z = {\rm aggr} ( ({\rm conv}(o)) \qquad \in \mathbb{R}^{1 \times d'}
    \label{eq:non-local}
\end{align}
where the aggr($\cdot$) function can be a pool operation, such as max-pooling and mean-pooling.

\subsection{GPNN for node classification}
With aforementioned designs, we build our proposed graph pointer neural networks for node classification tasks. We combine  the non-local features from non-local aggregation in Equation (\ref{eq:non-local}), neighborhood structural information from the node embedding step in Equation (\ref{eq:neiborhood}) and each node's ego-feature to obtain the final representation. A linear transformation $W\in \mathbb{R}^{F\times d_0} $ is performed on the ego-feature $x\in \mathbb{R}^{1\times F} $ and the three type of embeddings are combined via concatenation.
\begin{align}
    x_{final} = {\rm concat}(xW,\hat{x},z) \qquad \in \mathbb{R}^{1\times(d_0+d+d')}
\end{align}Finally, we utilize a fully-connected layer to make predictions of node classfication task and train the model with a cross-entropy loss.
\begin{align}
    &y_{pred} = softmax(FFN(x_{final}))\\
    &\mathcal{L} = \sum_{i=1}^{N_l}{y_i\log y_{pred\ i}}
\end{align}
where the number of labeled nodes is $N_l$.

\section{Experiments}
For a comprehensive evaluation of GPNN, we conduct experiments for node classification tasks and compare it with various baselines. Analyses are also performed to reveal GPNN's privilege in filtering out irrelevant neighbors and reducing over-smoothing.
\begin{table*}[htb]
    \renewcommand\arraystretch{1.2}
    \centering
    \small
    \scalebox{1.0}{
    \begin{tabular}{lcccccc}
    \toprule
    \textbf{Datasets} &\textbf{ Chameleon} & \textbf{Squirrel} & \textbf{Actor} & \textbf{Cornell} & \textbf{Texas} & \textbf{Wisconsin}\\
    \midrule
    \#Nodes & 2277 & 5201 & 7600 & 183 & 183 & 251\\
    \#Edges & 36101 & 217073 & 33544 & 295 & 309 & 499\\
    \#Features & 2325 & 2089 & 931 & 1703 & 1703 & 1703\\
    \#Classes & 5 & 5 & 5 & 5 & 5 & 5\\
    \#Homophily ratio $H(\mathcal{G})$ & 0.25 & 0.22 & 0.24 & 0.11 & 0.06 & 0.16\\
    \bottomrule
    \end{tabular}
    }
    \caption{Statistics and properties of benchmark datasets with heterophily.}
    \label{tab:datasets}
\end{table*}

\begin{table*}[htb]
    \renewcommand\arraystretch{1.2}
    \centering
    \small
    \scalebox{1.0}{
    \begin{tabular}{lccccccc}
    \toprule
    \textbf{Methods} & \textbf{Chameleon} & \textbf{Squirrel} & \textbf{Actor} & \textbf{Cornell} & \textbf{Texas} & \textbf{Wisconsin} & \textit{Average}\\
    \midrule
    MLP & 47.36{\tiny$\pm$2.37} & 29.82{\tiny$\pm$1.99} & 35.79{\tiny$\pm$1.09} & 82.16{\tiny$\pm$7.45} & 81.08{\tiny$\pm$3.82} & 85.49{\tiny$\pm$4.99} & 60.28  \\
    GCN~\cite{kipf2016semi} & 65.92{\tiny$\pm$2.58} & 49.78{\tiny$\pm$2.06} & 30.16{\tiny$\pm$1.27} & 58.91{\tiny$\pm$8.33} & 59.73{\tiny$\pm$3.24} & 58.82{\tiny$\pm$6.06} & 53.89  \\
    GAT~\cite{velivckovic2017graph} & 65.32{\tiny$\pm$2.00} & 46.79{\tiny$\pm$2.08} & 29.74{\tiny$\pm$1.46} & 56.76{\tiny$\pm$5.70} & 59.45{\tiny$\pm$6.37} & 57.06{\tiny$\pm$7.07} & 52.52 \\
    GraphSage~\cite{ying2018graph} & 58.73{\tiny$\pm$1.68} & 41.61{\tiny$\pm$0.74} & 34.23{\tiny$\pm$0.99} & 75.95{\tiny$\pm$5.01} & 82.43{\tiny$\pm$6.14} & 81.18{\tiny$\pm$5.56} & 62.36\\
    MixHop~\cite{abu2019mixhop} & 60.50{\tiny$\pm$2.53} & 43.80{\tiny$\pm$1.48} & 32.22{\tiny$\pm$2:34} & 73.51{\tiny$\pm$6.34} & 77.84{\tiny$\pm$7.73} & 75.88{\tiny$\pm$4.90}  & 60.58\\
    \midrule
    Geom-GCN~\cite{pei2020geom} & 60.90 & 38.14 & 31.63 & 60.81 & 67.57 & 64.12 & 53.86 \\
    H2GCN~\cite{zhu2020beyond} & 59.39{\tiny$\pm$1.98} & 37.90{\tiny$\pm$2.02} & 35.86{\tiny$\pm$1.03} & 82.16{\tiny$\pm$4.80} & 84.86{\tiny$\pm$6.77} & 86.67{\tiny$\pm$4.69} & 64.47 \\
    Node2Seq~\cite{yuan2021node2seq} & 69.4{\tiny$\pm$1.6} & 58.8{\tiny$\pm$1.4}  & 31.4{\tiny$\pm$1.0} & 58.7{\tiny$\pm$6.8} & 63.7{\tiny$\pm$6.1} & 60.3{\tiny$\pm$7.0} & 57.05 \\
    \midrule
    \textbf{GPNN} (ours) & \textbf{71.27}{\tiny$\pm$1.88} & \textbf{59.11}{\tiny$\pm$1.13} & \textbf{37.08}{\tiny$\pm$1.41} & \textbf{85.14}{\tiny$\pm$6.00} & \textbf{85.23}{\tiny$\pm$6.40} & \textbf{86.86}{\tiny$\pm$2.62} & \textbf{70.78} \\
    \bottomrule
    \end{tabular}
    }
    \caption{Mean accuracy$\pm$stdev over different data splits on the six real-world heterophilic graph datasets. The best result is highlighted.}
    \label{tab:experiment}
\end{table*}

\subsection{Datasets}
We evaluate our proposed graph pointer neural networks (GPNN) on six public heterophilic graph datasets. The dataset statistics are summarized in Table \ref{tab:datasets}.

\begin{itemize}
    \item \textbf{Chameleon and Squirrel} are subgraphs of web pages in Wikipedia~\cite{rozemberczki2021cham_squ}, where nodes represent web pages regarding corresponding topics, edges denote mutual links between pages, and node features correspond to several informative nouns in the Wikipedia pages. All nodes are classified into five categories based on the average monthly traffic of the web page.
    \item \textbf{Actor} is a subgraph extracted from film-director-actor-writer network ~\cite{tang2009actor}, where each node corresponds to an actor, edge between two nodes denotes co-occurrence on the same Wikipedia page, and node features correspond to some keywords in the Wikipedia pages. All nodes are classified into five categories according to the types of the actors.
    \item \textbf{Cornell, Texas and Wisconsin} are three subsets of the WebKB dataset collected by CMU, which represent links between web pages of the corresponding universities. In these networks, nodes represent web pages, edges are hyperlinks between them, and node features are the bag-of-words representation of web pages. All nodes are classified into five categories: student, project, course, staff, and faculty.
\end{itemize}

\subsection{Baselines}
\begin{itemize}
    \item  A simple MLP model that ignores the graph structure.
    \item Four traditional GNN models for node classification tasks: GCN ~\cite{kipf2016semi}, GAT~\cite{velivckovic2017graph}, GraphSage~\cite{ying2018graph} and MixHop~\cite{abu2019mixhop}.
    \item Three state-of-the-art models addressed heterophily: Geom-GCN~\cite{pei2020geom}, H2GCN~\cite{zhu2020beyond} and Node2Seq~\cite{yuan2021node2seq}.
\end{itemize}

\subsection{Experimental setup}
To make a fair comparison, the number of layers in MLP, GCN and GAT is set to 2. We run 2000 epochs and apply an early stopping strategy with a patience of 100 epochs on both the cross-entropy loss and accuracy on the validation set to choose the best model. For GPNN, the depth of node sampling is 2 with a max sequence length of 16. Other hyperparameters are tuned on the validation set: hidden unit $\in$ \{16, 32, 64\},  learning rate $\in$ \{0.01, 0.005\}, dropout in each layer $\in$ \{0, 0.5, 0.99\}, weight decay $\in$ \{1E-3, 5E-4, 5E-5, 5E-6\}, number of the selected nodes from each sequence $\in$ \{1, 2, 4, 8\}. Our methods are implemented using Pytorch and Pytorch Geometric. We closely follow the experimental procedure with ~\citet{zhu2020beyond}. For all datasets, we use the same feature vectors, labels and ten random splits provided by \citet{pei2020geom}.

\subsection{Results on heterophilic graphs}
Table \ref{tab:experiment} summarizes the prediction results of node classification for six datasets. We report the mean accuracy with a standard deviation over ten different data splits.

\paragraph{Comparision between MLP and GNNs.} As most nodes within the local neighborhood have different features or labels in heterophilic graphs, local aggregation brings more noises than helpful information. Even a simple MLP model outperforms classic GNNs on Actor, Cornell, Texas, and Wisconsin. GraphSage concatenates the ego-feature and neighborhood features explicitly, which shows better performance compared to those mix up different features (e.g., GCN and GAT) on these four datasets, too. This recalls our motivation that it is necessary to distinguish helpful nodes from massive neighbors. On the other hand, the local structural information is more critical for Chameleon and Squirrel, on which GNNs behave better than MLP. Nevertheless, as we will discuss later, ranking of relevant neighboring nodes is more beneficial for these two datasets.

\paragraph{Comparison between GPNN and SOTAs.} With selective non-local aggregation, GPNN achieves a better balance between neighborhood information and non-local semantic features. It significantly improves the performance over all six benchmarks, with an average lift of 6.3\% over the best state-of-the-art method. Especially, we improve 1.9\% on Chameleon and 3.0\% on Cornell respectively over the second-best results. As GPNN ignores most irrelevant nodes, it shows consistent improvements compared with H2GCN, which does not explicitly distinguish nodes from multi-hop neighborhoods. Node2Seq shows remarkable performances on both Chameleon and Squirrel; however, since the nodes' ranking is not jointly optimized, it is hard to select the most valuable nodes. In addition, mixing up the neighborhood nodes (see the results of Node2Seq) also reduces the performance of the other four datasets. GPNN considers three different kinds of information jointly, including ego-features, structural neighborhood, non-local semantic features, and adopts a practical nodes ranking module to filter out irrelevant nodes. This design consistently shows advantages in different scenarios.

\begin{figure}[t]
	\centering
        \includegraphics[width=\columnwidth]{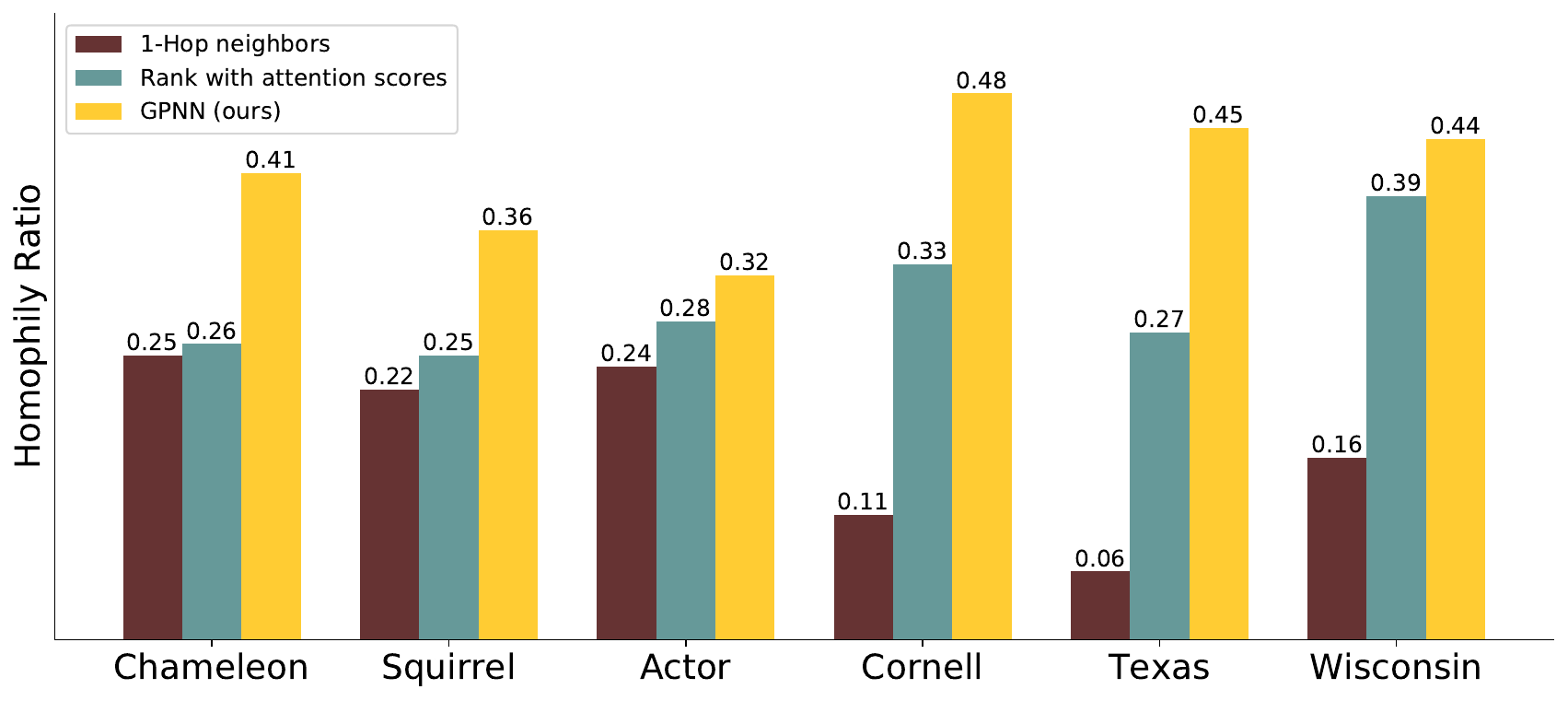}
        \caption{Comparision of homophily ratios between 1-hop neighbors, nodes ranked with attention scores and nodes selected with pointer network in GPNN.}
        \label{fig:homophily}
\end{figure}

\subsection{Analysis of node ranking}
Noises introduced by irrelevant nodes within the local neighborhood severely harm the classification results. Our motivation is that we can filter out the irrelevant nodes, i.e., improve the homophily ratio $H(\mathcal{G})$ of a graph. Then the aggregation would be more informative and accurate. Graph Pointer Neural Networks achieve this goal by learning an ordered sequence of most relevant neighboring nodes. Previous state-of-the-art, Node2Seq~\cite{yuan2021node2seq} applies attention-based approaches to calculate relevance scores and rank the neighboring nodes in descending order. However, the ranking procedure is non-differentiable and can not be optimized jointly in an end-to-end pipeline. 

Our analysis conducts experiments to demonstrate the advantages of ranking nodes through the graph pointer network in a joint-optimized manner. For each node, we find the most relevant five nodes ranked by graph pointer network and attention scores in Node2Seq, respectively, and then calculate the homophily ratio of these nodes. To compare with traditional GNNs, we also randomly select five 1-hop neighbors and calculate their homophily ratio for reference. The results are summarized in Figure \ref{fig:homophily}. We observe that the node ranking strategies in GPNN and Node2Seq both improve the homophily ratio of original 1-hop neighbors, indicating the effectiveness of neighborhood selection. Moreover, compared with Node2Seq, GPNN achieves better results, owing to its capability of joint optimization. Specifically, the average homophily ratio of GPNN is improved by 46\% over Node2Seq.

\subsection{Analysis of over-smoothing}
\begin{figure}[t]
	\centering
	\begin{subfigure}[t]{0.49\columnwidth}
         \centering
         \includegraphics[width=\columnwidth]{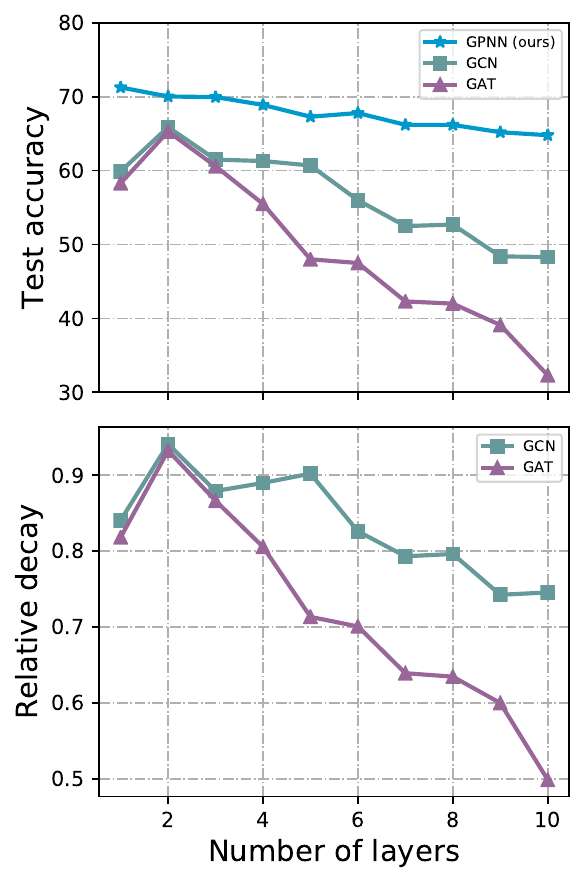}
         \caption{Chameleon}
     \end{subfigure}
     \hfill
     \begin{subfigure}[t]{0.49\columnwidth}
         \centering
         \includegraphics[width=\columnwidth]{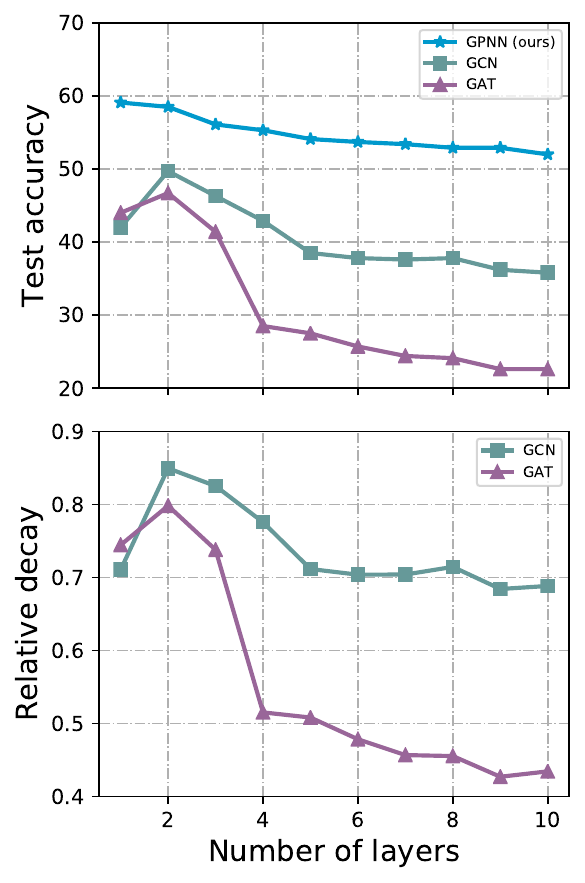}
         \caption{Squirrel}
     \end{subfigure}
       \caption{Over-Smoothing on Chameleon and Squirrel datasets. The top figures show the test accuracy while the bottom ones are relative decays of GCN and GAT compared to GPNN.}
       \label{fig:over-smoothing}
\end{figure}

Many GNNs suffer from the over-smoothing issue. When the layers of the GNN model increase, the mixture of neighborhood features by graph convolution tends to be indistinguishable, as important discriminating information from the input is erased~\cite{li2018deeper,dehmamy2019understanding}. We study the degree of over-smoothing by stacking a different number of GCN, GAT, GPNN layers and comparing their test accuracy of node classification. Figure \ref{fig:over-smoothing} shows the results on both Chameleon and Squirrel datasets. The top figures are the test accuracy of the three models with the different number of layers, and the bottom ones are relative decays compared to GPNN. For GCN and GAT, the test accuracy decreases fast when stacking more than two layers. For example, the test accuracy on Chameleon drops by 18\% and 33\% for GCN and GAT, respectively. For GPNN, the test accuracy decrease from 0.71 to 0.65, resulting in a slight decay of 6\%. This indicates that GPNN is more powerful to alleviate over-smoothing. 

\section{Conclusion}
In this work, we focus on the node representation learning of heterophilic graphs and present a novel GNN framework termed Graph Pointer Neural Networks (GPNN). Since most of the connected nodes in heterophilic graphs always possess dissimilar features or belong to different classes, we propose to incorporate a graph pointer generator to the GNN architecture, which distinguishes crucial information from distant nodes and performs non-local aggregation selectively. Experiments demonstrate the superiority of GPNN over previous state-of-the-art GNN models. In addition, extensive analyses are conducted to show GPNN's advantages in both filtering irrelevant nodes and alleviating over-smoothing phenomena. In future works, we will explore more techniques for improving the scalability of GPNN, such as advanced sampling strategies and more efficient network architectures.

\bibliography{aaai22}
\end{document}